\documentclass{article}

\usepackage[preprint,nonatbib]{neurips_2022}

\usepackage{multicol}
\usepackage{wrapfig}
\usepackage[utf8]{inputenc} 
\usepackage[T1]{fontenc}    
\usepackage{hyperref}       
\usepackage{url}            
\usepackage{booktabs}       
\usepackage{amsfonts}       
\usepackage{nicefrac}       
\usepackage{microtype}      
\usepackage{xcolor}         
\usepackage[normalem]{ulem}
\useunder{\uline}{\ul}{}
\usepackage{multirow}
\usepackage{caption}
\usepackage{array}
\newcolumntype{C}[1]{>{\centering\arraybackslash}p{#1}}
\usepackage{longtable} 
\usepackage{changepage}

\makeatletter
\usepackage{multirow}

\newcommand*{\rom}[1]{\expandafter\@slowromancap\romannumeral #1@}
\usepackage{graphicx,scalerel}

\newtheorem{innercustomgeneric}{\customgenericname}
\providecommand{\customgenericname}{}
\newcommand{\newcustomtheorem}[2]{%
  \newenvironment{#1}[1]
  {%
   \renewcommand\customgenericname{#2}%
   \renewcommand\theinnercustomgeneric{##1}%
   \innercustomgeneric
  }
  {\endinnercustomgeneric}
}

\newcustomtheorem{customthm}{Theorem}
\newcustomtheorem{customlemma}{Lemma}
\newcustomtheorem{customprop}{Proposition}
\newcustomtheorem{customdef}{Definition}
\newcustomtheorem{customcor}{Corollary}

\usepackage{amsmath,amsfonts,bm}

\def\1{\bm{1}}

\DeclareMathAlphabet{\mathsfit}{\encodingdefault}{\sfdefault}{m}{sl}
\SetMathAlphabet{\mathsfit}{bold}{\encodingdefault}{\sfdefault}{bx}{n}

\DeclareMathOperator*{\argmin}{arg\,min}

\usepackage{hyperref}
\usepackage{url}
\usepackage{amssymb}
\usepackage{enumerate}

\title{AutoDiff: combining {\ul Auto}-encoder and {\ul Diff}usion model for tabular data synthesizing}

\author{
  Namjoon Suh\\
  UCLA\\
  \texttt{namjsuh@ucla.edu}\\
  \And
  Xioafeng Lin \\
  UCLA \\
  \texttt{bernardo1998@g.ucla.edu}\\
  \AND
  Din-Yin Hsieh \\
  UCLA \\
  \texttt{darrenhsieh1205@g.ucla.edu} \\
  \And
  Merhdad Honarkhah \\
  Meta \\
  \texttt{merhdad@meta.com} \\
  \And
  Guang Cheng \\
  UCLA \\
  \texttt{guangcheng@stat.ucla.edu} \\
}

\begin{document}

\maketitle

\begin{abstract}
Diffusion model has become a main paradigm for synthetic data generation in many subfields of modern machine learning, including computer vision, language model, or speech synthesis. 
In this paper, we leverage the power of diffusion model for generating synthetic tabular data.
The heterogeneous features in tabular data have been main obstacles in tabular data synthesis, and we tackle this problem by employing the auto-encoder architecture. 
When compared with the state-of-the-art tabular synthesizers, the resulting synthetic tables from our model show nice statistical fidelities to the real data, and perform well in downstream tasks for machine learning utilities.
We conducted the experiments over $15$ publicly available datasets.  
Notably, our model adeptly captures the correlations among features, which has been a long-standing challenge in tabular data synthesis.
Our code is available at https://github.com/UCLA-Trustworthy-AI-Lab/AutoDiffusion.
\end{abstract}

\section{Introduction}
The creation of synthetic tabular data is invaluable for research, testing, and analysis, especially when real-world data is scarce or sensitive. It facilitates scenario exploration, algorithm testing, and practical data analysis experiences for students and professionals. Additionally, synthetic tabular data serves as a benchmark for evaluating data processing and predictive models, ensuring safe performance assessment. It addresses privacy, data scarcity, and accessibility issues, offering new possibilities for data-driven research in academia and industry.

Given the importance of synthesizing tabular data, many researchers have put enormous efforts on building tabular synthesizers with fidelity and utility guarantees. 
CTGAN~\cite{xu2019modeling} and its variants~\cite{zhao2021ctab,zhao2022ctab} (e.g., CTABGAN, CTABGAN+) have gained popularity for generating tabular data using a Generative Adversarial Networks~\cite{goodfellow2020generative} (GANs). 
These models employ advanced data encoders, modeling continuous variables with Gaussian Mixture models (GMM) and discrete variables with one-hot encoding. 
However, GMM may not work well for certain real-world continuous variables, and one-hot encoding can increase data dimensionality for discrete variables, requiring large neural networks. 

With the rise of Dalle-2~\cite{ramesh2022hierarchical}, diffusion models~\cite{song2020score} have excelled, outperforming Generative Adversarial Network (GAN)\cite{goodfellow2020generative} models in various domains like image synthesis\cite{dhariwal2021diffusion}, medical imaging~\cite{muller2022diffusion}, etc.
Recently, diffusion-based tabular synthesizers, like Stasy~\cite{kim2022stasy}, have shown promise by preprocessing data with min-max scaling and one-hot encoding, outperforming GAN-based methods in various tasks. 
Yet, score-based diffusion models~\cite{song2020score} weren't initially designed for heterogeneous features. 
Newer approaches, such as those using Doob's h-transform~\cite{liu2022learning}, TabDDPM~\cite{kotelnikov2022tabddpm}, and CoDi~\cite{lee2023codi}, aim to address this challenge by combining different diffusion models~\cite{song2020score,hoogeboom2022equivariant} or leveraging contrastive learning~\cite{schroff2015facenet} to co-evolve models for improved performance on heterogeneous data.
Following this line of research, we present a new tabular data synthesizer combining the ideas of auto-encoder and diffusion model referred as \textbf{AutoDiff}. 
In the next subsection, we specify the contributions of our model in the context of challenges of tabular synthesizing.

\subsection{Main Contributions}
In this subsection, we present three challenges of tabular data synthesizing and the main ideas  of our model \textbf{AutoDiff} to tackle those issues. 
To facilitate the understandings of AutoDiff model, the ideas are often highlighted by brief comparisons with the other state-of-the-art (SOTA) models

\underline{\textbf{Heterogeneous features}} are the most challenging issue for building a tabular synthesizer, as real-world tables frequently have numerical, discrete, or even mixed-type features.  
We are tackling this challenge by combining the ideas from auto-encoder and score-based diffusion model~\cite{song2020score}.
An autoencoder is a type of neural network architecture that learns to encode input data into latent representations and reconstruct them back to  the original input.
We leverage the power of autoencoder for learning the \textit{``continuous respresentations''} of the original heterogeneous features in the latent space. 
Then, the learned representations are fed to diffusion model to generate new latent  representations. 
The trained decoder translates the newly generated representations back to the form of original heterogeneous features.
This idea nicely takes the advantages of both models as auto-encoder can deal with the heterogeneous features and diffusion model has been showing the great performances for learning the distributions on continuous space.

\underline{\textbf{Mixed-type feature}} is a type of features commonly observed in many real-world tabular data, where it has both numerical and discrete components. 
These features are specifically difficult to capture as most of generative models only focus on either learning distributions on numerical or discrete domains. 
So it is not clear by simply combining the two different models which are separately designed for numerical and discrete variables. 
In this paper, we tackle this issue by creating a dummy variable which encodes the frequency of repeated values in a mixed-type feature.
This dummy variable is appended to the pre-processed table as an input to the autoencoder so that the latent representations of autoencoder include the information about the dummy variable. 
The new latent representations generated from diffusion model are decoded back through pre-trained decoder. 
The output of decoder has both mixed-type feature and corresponding dummy feature.
The information of these two features are combined together as the final output. 
More detailed descriptions are given in section~\ref{sec2}.
It should be noted the mixed-type encoder introduced in~\cite{zhao2021ctab} is different from ours in a sense that they directly encode the rows via one-hot-encoding and parameters in GMM. 
To showcase the performance of our idea, we present comparison plots of generated versus real data of a mixed-type variable over $6$ different models in Figure $1$.

\underline{\textbf{Correlations of features}} are important statistical objects that need to be  captured when building a tabular synthesizer.
Yet, due to the heterogeneous nature of tabular data, capturing these correlations is even more demanding than capturing correlations among purely numerical or discrete features.
AutoDiff naturally circumvents this challenge by its construction as diffusion model learns the joint distribution of latent representations in the \textit{continuous} space. 
As long as auto-encoder in AutoDiff gives good latent representations of rows in the table, 
it should be expected AutoDiff can nicely capture the correlations among features. 
We would like to emphasize this contribution by contrasting our idea with the SOTA diffusion based method, TabDDPM, where each categorical variables are considered as independent (i.e., they use separate forward diffusion process  for each categorical variable.) and the categorical and numerical variables are also modelled as independent since they used two different types of diffusion models for numerical and discrete variables, respectively.
Similar arguments can be made for GAN-based methods (i.e., CTGAN, CTABGAN+) as they separately employ the GMM for numerical features.
We perform comparative experiments of these models with ours on various real world datasets whose results are given in Table $1$ in section~\ref{exp}.

\begin{figure}[!t]
  \centering
  \includegraphics[width=0.75\textwidth]{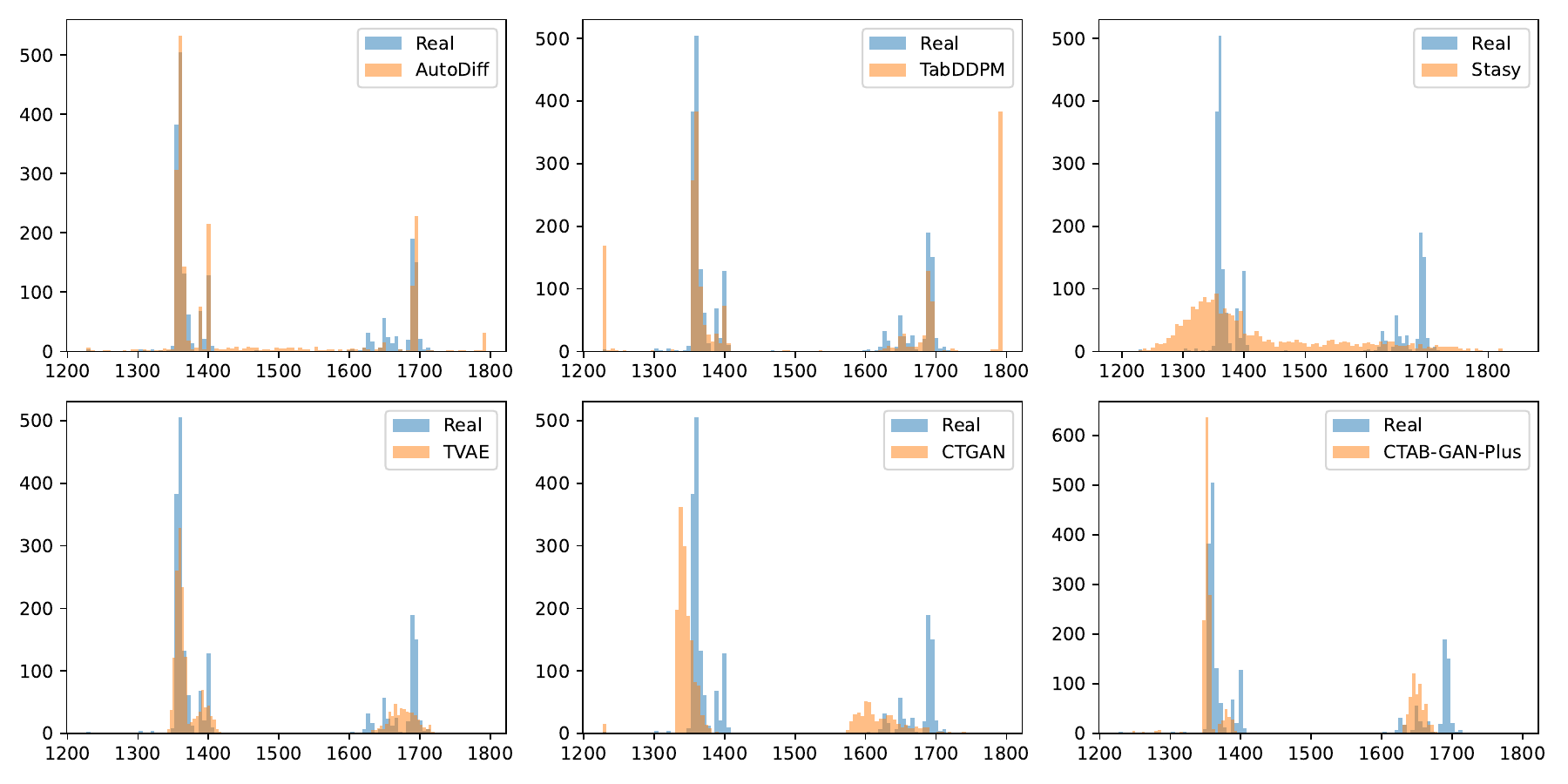}
  \caption{Comparison histograms of a generated and real ``Mixed-type'' feature over 6 models. The feature presented in the plot is \textit{``Length of Conveyer''} in faults dataset. } 
  \label{Fig1}
\end{figure}

\underline{\textbf{Numerical comparisons}} of AutoDiff with other models (with publicly available codes), namely,  CTGAN~\cite{xu2019modeling}, TVAE~\cite{xu2019modeling}, CTABGAN+~\cite{zhao2022ctab}, Stasy~\cite{kim2022stasy}, and TabDDPM~\cite{kotelnikov2022tabddpm} are conducted comprehensively across $15$ real-world datasets under various metrics.
In addition to the models mentioned above, we introduce the AutoGAN model, which is a customized adaptation of MedGAN~\cite{choi2017generating}. This inclusion is specifically tailored for the purpose of comparing diffusion model's performance with GAN model's under same auto-encoder.
Specifically, qualities of generated tables are measured through 
(1) statistical fidelities, 
(2) machine learning utilities in downstream tasks, and 
(3) privacy guarantees via Distances to Closest Records (DCR) in~\cite{zhao2021ctab}. 
\section{Proposed Method} \label{sec2}
In this section, we provide detailed descriptions on each component in our model.
Pre-and post-processing steps of the input and synthesized tabular data are introduced. 
Then, constructions on auto-encoder and diffusion models are provided.

\underline{\textbf{Pre- and post-processing steps.}} 
 It is essential to pre-process the real tabular data in a form that the machine learning model can extract the desired information from the data properly. 
We divide the heterogeneous features into three categories; (1) numerical, (2) discrete, and (3) mixed-type features.
Following is how we categorize the variables, and process each feature type. 
Let $\mathbf{x}$ be the column of a table to be processed.
\begin{enumerate}
    \item \textbf{\textit{Numerical feature}} : If $\mathbf{x}$'s entries are real-valued continuous, we categorize $\mathbf{x}$ as a numerical feature.
    Moreover, if the entries are integers with more than $25$ distinct values, e.g., ``Age in adult-income dataset'', the $\mathbf{x}$ is categorized as a numerical variable. 
    Here, $25$ is a user-specified threshold. 
    We will be using either min-max scaler or gaussian quantile transformation from scikit-learn library~\cite{pedregosa2011scikit} for pre-processing numerical features. 
    Hereafter, we denote $\mathbf{x}^{\text{Proc}}_{\text{Num}}$ as the processed column.
    
    \item \textbf{\textit{Discrete feature}} : If $\mathbf{x}$'s entries have string datatype, we categorize $\mathbf{x}$ as a discrete feature, e.g., ``Gender''.
    Additionally, the $\mathbf{x}$ with less than $25$ distinct integers is categorized as a discrete feature. 
    For pre-processing, we simply map the entries of $\mathbf{x}$ to the integers greater than or equal to $0$,
    and further divide the data type into two parts; binary and categorical, denoting them as $\mathbf{x}^{\text{Proc}}_{\text{Bin}}$ and $\mathbf{x}^{\text{Proc}}_{\text{Cat}}$. 

    \item \textbf{\textit{Mixed-type feature}} : Let's say $\mathbf{x}$ is categorized as a numerical feature. 
    If certain values in the $\mathbf{x}$ are repeated more than $h$-percentages of the entire data points, we consider the $\mathbf{x}$ as the mixed-type feature.
    Here, $h$ is a user-specified parameter, and we set it as $1$ throughout this paper. 
    The idea to encode the mixed-type variable is to create another discrete feature  variable $\mathbf{y}$, which encodes the labels of entries in~$\mathbf{x}$. 
    We label the entries repeated less than $h$-percent as $0$s.
    For entries repeated more than $h$-percent, we label them with the integers from $1$ to $K$ in the order of frequent repetition.
    Here, $K$ is the total number of entries which are repeated more than $h$-percent of the entire data points.

    \item \textbf{\textit{Post-processing step}} : After the AutoDiff model generates a synthetic dataset, it must be restored to its original format. For numerical features, this is achieved through inverse transformations, such as reversing min-max scaling or using inverse normal transformations. Integer labels in discrete features are mapped back to their original categorical or string values.    
    For mixed-type features, denote $\mathbf{y}^{\text{syn}}$ as the newly synthesized column from the AutoDiff model of the dummy variable $\mathbf{y}$ generated in the item $3$. 
    We combine the information of $\mathbf{x}^{\text{syn}}$ and $\mathbf{y}^{\text{syn}}$ leaving the entries in $\mathbf{x}^{\text{syn}}$ as they are, if the corresponding labels in $\mathbf{y}^{\text{syn}}$ are $0$.
    If the labels in $\mathbf{y}^{\text{syn}}$ are the integer values greater than $0$, then the corresponding entries in $\mathbf{x}^{\text{syn}}$ are replaced with the repeated values used to encode $\mathbf{y}$. 
\end{enumerate}

\begin{figure}[!t]
  \centering
  \includegraphics[width=0.75\textwidth]{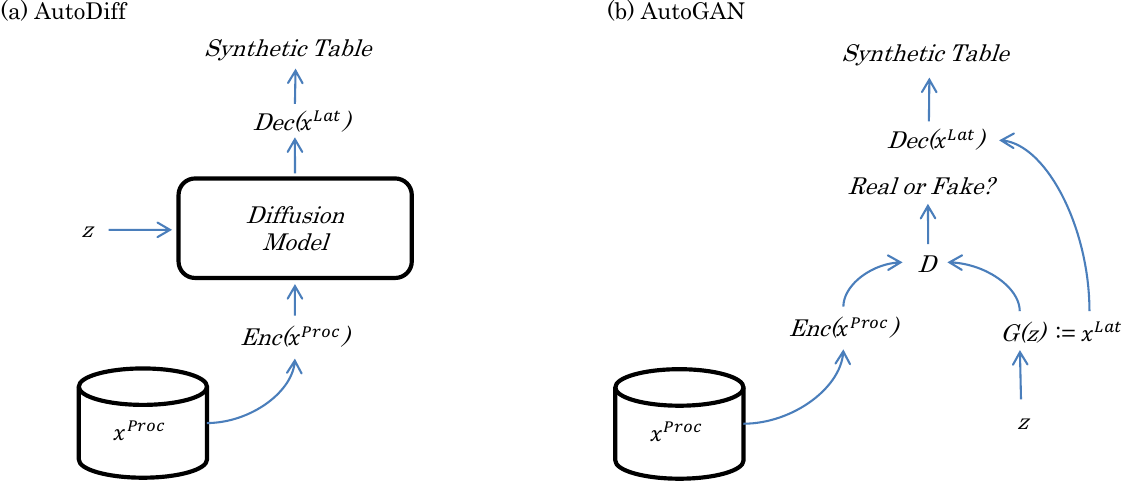}
  \caption{Structures of AutoDiff and AutoGAN models. Here $z$ are random inputs to the models for newly generated latent vectors $\mathbf{x}^{\text{Lat}}$. } 
  \label{Fig2}
\end{figure}

\underline{\textbf{Auto-encoder.}} 
The pre-processed input data $\mathbf{x}^{\text{Proc}}:=[\mathbf{x}^{\text{Proc}}_{\text{Num}};\mathbf{x}^{\text{Proc}}_{\text{Bin}};\mathbf{x}^{\text{Proc}}_{\text{Cat}}]$ is fed into the auto-encoder i.e., AE, and AE learns the latent representations of the input data. 
Network architecture of the AE simply consists of multilayer perceptron blocks with ReLU activation function. 
\begin{align}
    &\text{MLPBlock}(\mathbf{x}) = \text{ReLU}(\text{Linear}(\mathbf{x})) \nonumber \\
    &\mathbf{x}^{\text{Lat}} = \text{Linear}(\text{MLPBlock}(\cdots\text{MLPBlock}(\mathbf{x}^{\text{Proc}})))
    \label{Encoder} \\
    &\mathbf{x}^{\text{Out}} = \text{Linear}(\text{MLPBlock}(\cdots\text{MLPBlock}(\mathbf{x}^{\text{Lat}}))) \label{Decoder}
\end{align}
Let $\mathbf{x}^{\text{Out}}_{\text{Num}}$, $\mathbf{x}^{\text{Out}}_{\text{Bin}}$, and $\mathbf{x}^{\text{Out}}_{\text{Cat}}$ be the numerical, binary, and categorical outputs of the decoder~\eqref{Decoder}. 
Note that the dimensions of $\mathbf{x}^{\text{Out}}_{\text{Num}}$ and $\mathbf{x}^{\text{Out}}_{\text{Bin}}$ 
are same with their corresponding inputs, but the dimension of $\mathbf{x}^{\text{Out}}_{\text{Cat}}$ is $\sum_{i}K_{i}$ where $K_{i}$ is the number of categories in each categorical variable.
We set the dimensions of output features in this way as we used the mean-squared (MSE), binary cross entropy (BCE), and cross entropy (CE) in Pytorch package. 
Additionally, in order to keep the domain of the original numerical features, (e.g., ``Age'' cannot be negative.) we truncate the output of $\mathbf{x}^{\text{Out}}_{\text{Num}}$ with min and max values of $\mathbf{x}^{\text{Proc}}_{\text{Num}}$.
Finally, we minimize the following objective function:
\begin{align} \label{loss}
    \text{MSE}(\mathbf{x}^{\text{Proc}}_{\text{Num}},\mathbf{x}^{\text{Out}}_{\text{Num}})+
    \text{BCE}(\mathbf{x}^{\text{Proc}}_{\text{Bin}},\mathbf{x}^{\text{Out}}_{\text{Bin}})+
    \text{CE}(\mathbf{x}^{\text{Proc}}_{\text{Cat}},\mathbf{x}^{\text{Out}}_{\text{Cat}}).
\end{align}
It should be noted that the dimension of $\mathbf{x}^{\text{Lat}}$ is kept same with that of $\mathbf{x}^{\text{Proc}}$ and the outputs of all the other layers are set as $250$ throughout the paper.

\noindent\underline{\textbf{Stasy- \& Tab-AutoDiff.}}
The score-based diffusion model~\cite{song2020score} is utilized to generate new latent representations of $\mathbf{x}^{\text{Lat}}$.
Due to space limitations, we defer the detailed descriptions of the diffusion model to Appendix~\ref{Appendix2}.
In this work, we utilize Variance Preserving (VP)-SDE for data perturbation, and Euler-Maruyama method~\cite{song2020score} for sampling over the experiments.
Here, we devise two types of AutoDiff models. 
\textbf{Stasy-AutoDiff} (resp. \textbf{Tab-AutoDiff}) adopts the same time-dependent score network used in Stasy~\cite{kim2022stasy} (resp. TabDDPM~\cite{kotelnikov2022tabddpm}) and min-max scaler (resp. Gaussian Quantile Transformation) is used for pre-processing numerical variables.
The constructions of the score networks with detailed architectural settings for two models are deferred to Appendix~\ref{Appendix2}.
The reason for employing the same score networks is to facilitate comparisons of the performances between the Stasy-AutoDiff (resp. Tab-AutoDiff) and Stasy (resp. TabDDPM).
(The n\"aive form of Stasy without fine-tuning step is employed in this work.)
The overall pipeline of AutoDiff is given in Figure~\ref{Fig2}.

\noindent\underline{\textbf{Med-AutoDiff \& AutoGAN.}}
MedGAN~\cite{choi2017generating} has a very similar structure with AutoDiff in a sense that they used the idea of combining auto-encoder and GAN model. 
The model is only applicable to datasets with discrete variables.
Notably, they used MSE for categorical variables instead of CE-loss in~\eqref{loss}.
Following this set-up, we modify the loss~\eqref{loss} and refer the trained model as Med-AutoDiff.  
Furthermore, we design AutoGAN model which is applicable to heterogeneous features. 
We take the advantage of auto-encoder mentioned above, and used the same generator and discriminator in GAN model introduced in MedGAN~\cite{choi2017generating}.
Specifically, we use the minibatch averaging for discriminator and batch normalization and skip connection for generator. Same network architectures are used for the constructions of networks as in~\cite{choi2017generating}.
The most notable difference between AutoGAN and MedGAN is from the arguments taken in discriminator: in AutoGAN, discriminator distinguishes the real from fake \textbf{latent} vectors, whereas in MedGAN, it distinguishes the real from fake \textbf{decoded} latent vectors, and finetune the decoder while training discriminator. 
We choose to fix the parameters in decoder for the fair comparisons with Med-AutoDiff. 
See Figure~\ref{Fig2}.
Also note that we use the same auto-encoder loss of Med-AutoDiff in AutoGAN.

\section{Numerical Experiments \& Future Works} \label{exp}
\underline{\textbf{Datasets.}} We collect $15$ different real datasets to run the quantitative comparison experiments of AutoDiff model. 
These datasets are commonly employed to test the performances of recent tabular data synthesizers.
Links for the datasets are provided in Appendix~\ref{Appendix1}. 
Across all 6 models, 10 synthetic tables for each of the 15 datasets are generated.

\begin{table}[t!] \label{tab1}
\resizebox{\columnwidth}{!}{
\begin{tabular}{l|cc|ccc}
\multicolumn{1}{c|}{\multirow{2}{*}{\textbf{Models}}} & \multicolumn{2}{c|}{\textbf{Marginal}}                                                                                                                          & \multicolumn{3}{c}{\textbf{Joint}}                                                                                                                                                                                                            \\ \cline{2-6} 
\multicolumn{1}{c|}{}                                 & \multicolumn{1}{c|}{\begin{tabular}[c]{@{}c@{}}Num Dist.\\ (Wasserstein)\end{tabular}} & \begin{tabular}[c]{@{}c@{}}Cate Dist.\\ (Jensen Shannon)\end{tabular} & \multicolumn{1}{c|}{\begin{tabular}[c]{@{}c@{}}Num-Num\\ (Pearson Corr)\end{tabular}} & \multicolumn{1}{c|}{\begin{tabular}[c]{@{}c@{}}Cat-Cat\\ (Theil's U)\end{tabular}} & \begin{tabular}[c]{@{}c@{}}Num-Cat\\ (Correl Ratio)\end{tabular} \\ \hline
CTGAN                                                 & \multicolumn{1}{c|}{5.10}                                                               & \textbf{3.86}                                                         & \multicolumn{1}{c|}{1.60}                                                             & \multicolumn{1}{c|}{0.27}                                                          & 0.35                                                             \\
TVAE                                                  & \multicolumn{1}{c|}{4.05}                                                         & 5.77                                                                  & \multicolumn{1}{c|}{0.93}                                                       & \multicolumn{1}{c|}{0.70}                                                          & {\ul 0.30}                                                       \\
CTABGAN+                                              & \multicolumn{1}{c|}{4.63}                                                               & 4.61                                                                  & \multicolumn{1}{c|}{1.59}                                                             & \multicolumn{1}{c|}{0.39}                                                          & 0.53                                                             \\
Stasy                                                 & \multicolumn{1}{c|}{7.95}                                                               & 5.23                                                            & \multicolumn{1}{c|}{2.88}                                                             & \multicolumn{1}{c|}{{\ul 0.26}}                                                    & 0.56                                                             \\
TabDDPM                                               & \multicolumn{1}{c|}{4.42}                                                               & 5.11                                                                  & \multicolumn{1}{c|}{2.16}                                                             & \multicolumn{1}{c|}{0.35}                                                          & 0.60                                                             \\ 
AutoGAN                                               & \multicolumn{1}{c|}{6.49}                                                               & 5.23                                                                  & \multicolumn{1}{c|}{0.78}                                                             & \multicolumn{1}{c|}{1.65}                                                          & 0.62                                                             \\ \hline
Stasy-AutoDiff                                        & \multicolumn{1}{c|}{\textbf{3.57}}                                                      & {{\ul 4.42}}                                                                  & \multicolumn{1}{c|}{{\ul 0.66}}                                                    & \multicolumn{1}{c|}{\textbf{0.15}}                                                 & \textbf{0.15}    \\
Med-AutoDiff                                        & \multicolumn{1}{c|}{{\ul 3.67}}                                                      & 5.32                                                                  & \multicolumn{1}{c|}{\textbf{0.46}}                                                    & \multicolumn{1}{c|}{0.69}                                                 & 0.37   \\
Tab-AutoDiff                                        & \multicolumn{1}{c|}{3.92}                                                      & 5.32                                                                  & \multicolumn{1}{c|}{0.81}                                                    & \multicolumn{1}{c|}{\textbf{0.15}}                                                 & 0.17  
\end{tabular}}
\caption{Columns $1,2$ : Fidelity ranks of marginal distributions over all datasets in terms of Wasserstein distance between numerical columns and Jensen-Shannon divergence between categorical columns.
Columns $3, 4, 5$ : $L^{2}$-differences between correlation matrices computed on real and synthesized tables under three metrics.}
\vspace{-15pt}
\end{table}

\underline{\textbf{Fidelity Comparisons.}} 
Fidelity of the synthesized table refers to how well the generated table retains the statistical properties of the original dataset. 
Inspired from~\cite{zhao2021ctab,kotelnikov2022tabddpm}, we assess the similarities in marginal and joint feature distributions between the real (\textbf{R}) and synthetic (\textbf{S}) tables using a range of metrics. Specifically, for numerical features, we employ the Wasserstein distance (\textbf{WD}) to quantify the differences between \textbf{R} and \textbf{S}. For categorical features, we use Jensen-Shannon (\textbf{JS}) divergence to gauge the similarity in their distributions. 
The first two columns in Table $1$ record the averaged rankings of models for the respective dissimilarity measures applied to the columns between the \textbf{R} and \textbf{S} across all $150$ datasets considered for evaluation. 
Additionally, we evaluate joint distribution quality using various methods depending on the feature types. This includes using the Pearson correlation coefficient for numerical-numerical feature relationships, Theil's U statistics for assessing the dependence between categorical-categorical features, and the correlation ratio for categorical-numerical feature associations. 
The last three columns in Table $1$ record the averaged $L^{2}$-distance of computed correlation matrices between $\mathbf{R}$ and $\mathbf{S}$ across $150$ synthesized tables for each model.
The result shows that the three AutoDiff models show either the best results (bold) or the second best results (underlined) for all $5$ metrics.
It turns out that the AutoDiff recovers the numerical features pretty well, but performs relatively poor on the recovery of categorical features, when compared with CTGAN. 
The good performances on \textbf{WD} metric is due from the combination of  sophisticated design of auto-encoder, specifically for the mixed-type features, and the power of diffusion model for its ability to learn continuous variables. 
The role of the auto-encoder is even manifested when the result is compared with Stasy and TabDDPM. 
Recall that exactly the same diffusion models with Stasy-AutoDiff and Tab-AutoDiff are employed in Stasy and TabDDPM.
Turning our attentions to correlation measures, three AutoDiff models shows the best performances for the three metrics as presented in the Table $1$. 
This matches with our expectations as diffusion model in AutoDiff is designed to capture the joint distribution of $\mathbf{x}^{\text{Lat}}$ unlike other models where they employ different models for different data types. 
Stasy turns out to perform well at capturing the correlations among categorical variables, and TVAE is good at capturing Num-Cat correlations. 
Overall AutoDiff outperforms AutoGAN model in fidelity measures for all $5$ metrics.
We present the correlation plots of AutoDiff and TabDDPM for Abalone dataset in the Appendix~\ref{Appendix3}. 

\underline{\textbf{Utility Comparisons.}} 
We evaluated the efficacy of AutoDiff in preserving the Machine Learning Utility of synthesized data, following the Train on Synthetic and Test on Real (TSTR)~\cite{esteban2017real} pipeline. 
For each real dataset, we split it into training/test set with the ratio of 80\%/20\%. 
Then we fit two prediction models, one on the training set of real data and one on the corresponding synthetic table. Finally we test the performance of both models on the test set of the real data. For classification tasks we used Na\"ive Bayes, K-Nearest Neighbors, Decision Tree, Random Forest, XGBoost, LightGBM, and CatBoost. For regression tasks we used Linear Regression, Lasso regression, Decision Tree, Random Forest.
Table $2$ illustrates the aggregate performance metrics averaged across all datasets, the performances of classifiers and regressors trained on synthetic tables generated by various synthesizers are reported. Notably, AutoDiff consistently outperforms GAN and Autoencoder-based methods, such as CTGAN, CTABGAN+, AutoGAN, and TVAE by a substantial margin. When compared to other diffusion-based methods, AutoDiff exhibits superior performance; it notably surpasses the original Stasy method that does not employ autoencoder-based encoding. While the improvement that AutoDiff demonstrates over the DDPM-based model (TabDDPM) is on-par for classification tasks, it is significantly more pronounced for regression tasks, as evidenced by notably higher \( R^2 \) values and lower RMSE scores.
It is interesting to note the employment of gaussian quantile transformation and score-network structure from TabDDPM improves the performances in classification tasks when comparing Stasy- and Tab-Autodiffs over all three metrics.

\begin{table}[t!] \label{tab:2}
\centering
\begin{tabular}{l|C{1.8cm}|C{1.8cm}|C{1.8cm}|C{1.8cm}|C{2cm}}
\multirow{2}{*}{\textbf{Models}} & \multicolumn{3}{c|}{\textbf{Classification}}        & \multicolumn{2}{c}{\textbf{Regression}} \\ \cline{2-6} 
                                   & \textbf{Accuracy} & \textbf{F1} & \textbf{AUROC} & \textbf{$R^2$}     & \textbf{RMSE}    \\ \cline{1-6} 
  Identity           & {0.923}          & {0.809}   & {0.941}            & {0.633}            & {1817.307}        \\ \hline
 CTGAN                             & {0.699}      & {0.465}      & {0.742}            & {0.012}            & {8052.081}        \\
TVAE                              & {0.794}      & {0.616}      & {0.819}            & {\ul 0.18}            & {4699.311}        \\
CTABGAN+                           & {0.741}      & {0.482}      & {0.713}            & {0.061}            & {5518.107}                \\ 
StaSy                             & {0.793}      & {0.602}      & {0.808}            & { $\ll 0$ }            & {$1.06 \times 10^{8}$}        \\
TabDDPM                           & {0.820}   & {\textbf{0.665}}      & {\textbf{0.852}}            &{-9.39}            & {17322.613}        \\ 
AutoGAN                    & {0.618}   & {0.356}   & {0.647}       & {$\ll 0$}            & {$2.47 \times 10^{11}$} \\ \cline{1-6}
Stasy-AutoDiff                    & {{\ul 0.834}}   & { 0.620}   & { \ul 0.846}       & \textbf{0.199}            & {{\ul 4429.56}}  \\    
Med-AutoDiff                    & {0.777}   & {0.581}   & {0.819}       & {0.1}            & {\textbf{4281.12}} \\
Tab-AutoDiff                  & \textbf{{0.836}}   & {\ul 0.659}   & {\textbf{0.852}}       & {0.172}            & {{ 4679.11}} 
\end{tabular}
\vspace{1pt}
\caption{Classification/Regression Performance on Different Synthetic Data. We reported the average of accuracy, F1 score, AUROC for classification tasks, and $R^2$ and RMSE for regression tasks. Identity refers to the classifiers/regressors trained on the training set of the real data. Accuracy refers to the proportion of predictions that equal the real label.}
\vspace{-15pt}
\end{table}

\begin{wraptable}{r}{0.35\textwidth}
\vspace{-10pt}
\begin{tabular}{lcc} 
\toprule
\textbf{Models} & \textbf{MDCR Rank} \\
\midrule
CTGAN & 4.33  \\
TVAE & {\ul 7.00} \\
CTABGAN+ & 3.87 \\
Stasy & 1.93 \\
TabDDPM & 3.40 \\
AutoGAN & 4.93 \\ \hline
Stasy-AutoDiff & {6.13} \\
Med-AutoDiff & {\textbf{7.07}} \\
Tab-AutoDiff & 6.33 \\
\bottomrule
\end{tabular}
\caption{Rank of MDCR over different models. The highest rank (resp. second to the highest) is bolded (resp. underlined) implying  synthesized datasets are close to the real. } 
\label{tab3}
\vspace{-10pt}
\end{wraptable}

\underline{\textbf{Privacy Comparisons.}}
Privacy is an important issue in synthetic tabular data generation, and we simply measure it through the Distances to Closest Records (DCR) in~\cite{zhao2021ctab}.
For each synthetic sample, DCR computes the minimum L2-distance to real records, and Mean DCR averages these distances over all synthetic samples. 
Essentially, low DCR indicates synthetic samples memorize some data points in real table, violating some privacy restrictions. 
In contrast, high DCR values denote the synthesizer generates some new data points, which cannot be observed in real-table.
But note that some random noises can have high DCR values.
Therefore, DCR should be considered with fidelity and utility together.
In Table~\ref{tab3}, the averaged rankings of Mean-DCR (MDCR) for $9$ models are presented. 
High ranking indicates the model has low MDCR values. 
The reason why we adopt the ranking is due to the large variability of MDCR per dataset. 
We present the MDCR values per each dataset over $9$ models in the Appendix~\ref{Appendix4}.
The auto-encoder based methods, i.e., AutoDiff, AutoGAN, TVAE, have relatively higher ranks of MDCR, than non auto-encoder based methods such as Stasy or TabDDPM.
This result is expected as sophisticatedly designed auto-encoder frequently overfits the input data, combined with the memorization issues of diffusion model~\cite{carlini2023extracting}.
We observe that the overall rank of MDCR is similar to that of fidelity for numerical variables in Table $1$, and strongly conjecture the relatively good performance of TabDPPM in MDCR measure over AutoDiff is attributed to the fact that it does not capture the correlation structures among features but shows good fidelity to individual feature. 

\underline{\textbf{Future works.}} 
We believe ensuring the privacy guarantees with good statistical fidelity and utility of synthesized tabular data is one of the most fundamental questions we should resolve. Under a rigorous framework of differential privacy (DP)~\cite{dwork2006differential}, further extending the current form of AutoDiff model that satisfies the DP-constraints is the most promising directions for the future work~\cite{dockhorn2022differentially,lyu2023differentially}.



\newpage
\appendix
\section{Real Tabular-data List} \label{Appendix1}
We provide the URL for the sources of each dataset considered in the paper. 
The provided information also includes the downstream task that each dataset is designed for, which may include binary classification, multi-class classification, or regression.

\begin{enumerate}
    \item \textbf{abalone} (OpenML) : https://www.openml.org/search?type=data\&sort=runs\&id=183\&status=\\active (Multi class)
    \item \textbf{adult} (Kohavi, R) : Check the reference~\cite{kohavi1996scaling}. (Binary class)
    \item \textbf{Bean} (UCI) : https://archive.ics.uci.edu/dataset/602/dry+bean+dataset (Multi class)
    \item \textbf{Churn-Modelling} (Kaggle) : https://www.kaggle.com/datasets/shrutimechlearn/churn-modelling (Binary class)
    \item \textbf{faults} (UCI) : https://archive.ics.uci.edu/dataset/198/steel+plates+faults (Multi class)
    \item \textbf{HTRU} (UCI) : https://archive.ics.uci.edu/dataset/372/htru2 (Binary class)
    \item \textbf{indian liver patient} (Kaggle) : https://www.kaggle.com/datasets/uciml/indian-liver-patient-records?resource=download (Binary class)
    \item \textbf{insurance} (Kaggle) : https://www.kaggle.com/datasets/mirichoi0218/insurance (Regression)
    \item \textbf{Magic} (Kaggle) : https://www.kaggle.com/datasets/abhinand05/magic-gamma-telescope-dataset?resource=download (Binary class)
    \item \textbf{News} (UCI) : https://archive.ics.uci.edu/dataset/332/online+news+popularity (Regression)
    \item \textbf{nursery} (Kaggle) : https://www.kaggle.com/datasets/heitornunes/nursery (Multi class)
    \item \textbf{Obesity} (Kaggle) : https://www.kaggle.com/datasets/tathagatbanerjee/obesity-dataset-uci-ml (Multi class)
    \item \textbf{Shoppers} (Kaggle) : https://www.kaggle.com/datasets/henrysue/online-shoppers-intention (Binary class)
    \item \textbf{Titanic} (Kaggle) : https://www.kaggle.com/c/titanic/data (Multi class)
    \item \textbf{wilt} (OpenML) : https://www.openml.org/search?type=data\&sort=runs\&id=40983\&status=\\active (Binary class)
\end{enumerate}

\section{Score-based diffusion model} \label{Appendix2}
Diffusion model consists of two processes.
The first step, \textit{forward process}, transforms data into noise. 
Specifically, the score-based diffusion model~\cite{song2020score} uses the following stochastic differential equations (SDEs)~\cite{sarkka2019applied} for data perturbation:
\begin{align} \label{SDE}
    d\mathbf{x} = f(\mathbf{x}, t) dt + g(t) dW_{t},
\end{align}
where $f(\mathbf{x},t)$ and $g(t)$ are drift and diffusion coefficient, respectively, and 
$W_{t}$ is a standard Wiener-process (a.k.a. Brownian Motion) indexed by time $t\in[0,T]$. 
Here, the $f$ and $g$ functions are user-specified, and~\cite{song2020score} suggests three different types of SDEs, i.e., variance exploding (VE), variance preserving (VP), and sub-variance preserving (sub-VP) SDEs for data perturbation. 
In this paper, we used VP-SDE defined as : 
\begin{align} \label{VP-SDE}
    d\mathbf{x} = -\frac{1}{2}\beta(t) \mathbf{x} dt + \sqrt{\beta(t)} dW_{t},
\end{align}
where $\beta(t):=\beta(0) + (\beta(1)-\beta(0)) t $ with $\beta(1):=20$ and $\beta(0):=0.1$.
\\ \\
The second step, \textit{reverse process}, is a generative process that reverses the effect of the forward process. 
This process learns to transform the noise back into data by reversing SDEs in~\eqref{SDE}.
Through the Fokker-Planck equation of marginal density $p_{t}(\mathbf{x})$ of~\eqref{SDE} for each fixed time $t\in[0,T]$, the following reverse SDE~\cite{anderson1982reverse} can be easily derived:
\begin{align}\label{Reverse_SDE}
    d\mathbf{x} = \bigg[ f(\mathbf{x},t) - g(t)^{2} \nabla_{\mathbf{x}} \log p_{t}(\mathbf{x}) \bigg] dt + g(t) d\bar{W}_{t}.
\end{align}
Here, the gradient of $\log p_{t}(\mathbf{x})$ w.r.t to the perturbed data $\mathbf{x}(t)$ is referred to as score function, $dt$ in~\eqref{Reverse_SDE} is an infinitesimal negative time step, and $d\bar{W}_{t}$ is a Wiener-Process running backward in time, i.e., $t:T\rightarrow 0$. \
\\ \\
The score function, $\nabla_{\mathbf{x}} \log p_{t}(\mathbf{x})$, is approximated by a time-dependent score-based model $\mathbf{S}_{\theta}(\mathbf{x}(t),t)$ which is parametrized by neural networks in practice. 
The network parameter $\theta$ is estimated through minimizing the score-matching loss:
\begin{align} \label{sml}
    \theta^{\star}:=
    \argmin_{\theta}\mathbb{E}_{t\sim\mathcal{U}[\varepsilon,T]}\mathbb{E}_{\mathbf{x(t)\mid\mathbf{x(0)}}}\mathbb{E}_{\mathbf{x}(0)}\bigg[\lambda(t)^{2}\left\| \mathbf{S}_{\theta}(\mathbf{x}(t),t)-\nabla_{\mathbf{x}}\log p_{t}(\mathbf{x}(t)\mid\mathbf{x}(0)) \right\|_{2}^{2} \bigg],
\end{align}
where $\mathcal{U}[\varepsilon,T]$ is an uniform distribution over $[\varepsilon,T]$, and $\lambda(t)(> 0)$ is a positive weighting function that helps the scales of matching-losses $\left\| \mathbf{S}_{\theta}(\mathbf{x}(t),t) -\nabla_{\mathbf{x}}\log p_{0t}(\mathbf{x}(t)\mid\mathbf{x}(0)) \right\|_{2}^{2}$ to be in the same order across over the time $t\in[\varepsilon,T]$. 
The $\varepsilon$ is set up to prevent the divergence of score function when $t$ approached $0$, specifically when ground-truth $p_{0}(\mathbf{x})$ is non-smooth.
We set $\varepsilon:=1e^{-5}$ and $T=1$ throughout the paper.
The weight function is often chosen as $\lambda(t)\propto 1/\sqrt{\mathbb{E}\|\big[ \nabla_{\mathbf{x}}\log p_{0t}(\mathbf{x}(t)\mid\mathbf{x}(0))\big]\|_{2}^{2}}$. 
We set $\lambda(t):=1/\sigma(t)$, where $\sigma(t)$ is a standard deviation of $\mathbf{x}(t)\mid\mathbf{x}(0)$: $\sigma(t):=1-\exp\big(-\frac{1}{2}\int_{0}^{t}\beta(s)ds\big)$.
See Appendix B in~\cite{song2020score}.
\\ \\
\underline{\textbf{Score network of Stasy-AutoDiff.}}
Let $d$ be the dimension of $\mathbf{x}^{\text{Lat}}$ and the model is run over $t\in[10^{-5},1]$.
Under this setting, the score network is constructed as follows:
\begin{align}
    &\mathbf{h}_{0}=\mathbf{x}^{\text{Lat}}(t) \in \mathbb{R}^{d}, \nonumber \\
    &\mathbf{h}_{i}=\text{ELU}(\mathbf{H}_{i}(\mathbf{h}_{i-1},t) \oplus \mathbf{h}_{i-1}), \quad 1\leq i \leq d_{N}, \nonumber \\
    &S_{\mathbf{\theta}}(\mathbf{x}^{\text{Lat}}(t),t) = \text{Linear}(\mathbf{h}_{d_{N}}) \in \mathbb{R}^{d}, \nonumber
\end{align}
where $\mathbf{H}_{i}(\mathbf{h}_{i-1},t)$ is the concat-squash function~\cite{grathwohl2018ffjord} defined as 
$$
\mathbf{H}_{i}(\mathbf{h}_{i-1},t):=\text{Linear}_{i}(\mathbf{h}_{i-1}) \odot \phi(\text{Linear}_{i}^{\text{gate}}(t)+\text{Linear}_{i}^{\text{bias}}(t)).
$$
Here, $\odot$ is an element-wise multiplication, $\oplus$ means the concatenation operator, and $\phi$ is a sigmoid function.
Throughout this paper, we set $(d_{1},d_{2},d_{3},d_{4},d_{5})=(256, 512, 1024, 512, 256)$.
\\ \\
\underline{\textbf{Score network of Tab-AutoDiff.}} 
Denote $t\in[10^{-5},1]$ as a timestep, and SinTimeEmb refers to a sinusoidal time embedding as in~\cite{nichol2021improved} with dimension of $128$.
For any fixed $t\in[10^{-5},1]$, time embedding \textit{t-emb} and the input of the score-network $\mathbf{x}(t)$ is given as : 
\begin{align}
    &\textit{t-emb} = \text{Linear}(\text{SiLU}(\text{Linear}(\text{SinTimeEmb(t)}))),  \nonumber \\
    &\textbf{x}(t) = \text{Linear}(\mathbf{x}^{\text{Lat}}(t)) + \textit{t-emb}. \nonumber
\end{align}
All Linear layers in (11) and (12) have a fixed projection dimension $128$.
Then, we have the final output of time-dependent score network can be computed as follows: 
\begin{align}
    &\text{MLPBlock}(\mathbf{x}) = \text{Dropout}(\text{ReLU}(\text{Linear}(\mathbf{x}))), \nonumber \\
    &S_{\mathbf{\theta}}(\mathbf{x}^{\text{Lat}}(t),t) = \text{Linear}(\text{MLPBlock}(\cdots\text{MLPBlock}(\mathbf{x}(t)))). \nonumber 
\end{align}
In this paper, we used $4$ blocks of MLP.
\\ \\
Note that in Stasy-AutoDiff, min-max scaler is used for processing the numerical variables, whereas in Tab-AutoDiff, Gaussian Quantile Transformation is used for the processing of numerical variables.
\newpage
\section{Correlation Plots of Abalone} \label{Appendix3}
We provide the correlation plots of abalone dataset of synthetic tables from Stasy-AutoDiff and TabDDPM. 
The heatmaps on the top of Figures $3$ and $4$ display the difference of correlation plots from real and synthetic datasets.
We provide more correlation plots from models considered in the paper in 
\begin{figure}[!b]
  \centering
  \includegraphics[width=0.64\textwidth]{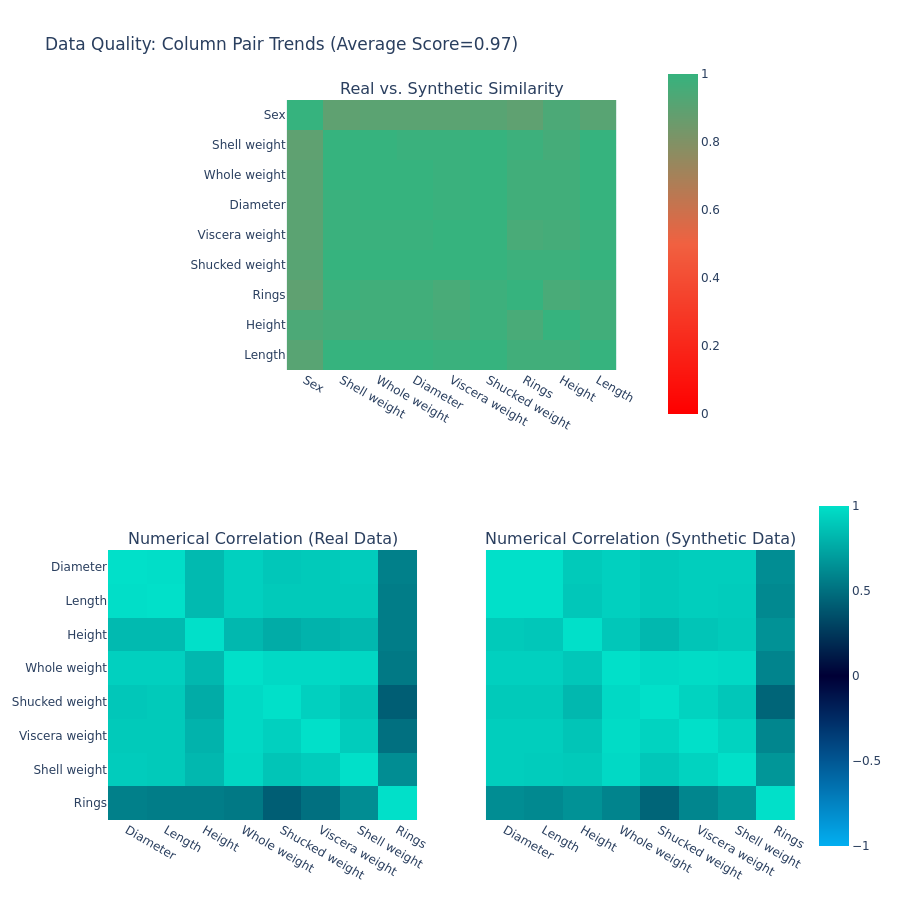}
  \caption{Correaltion Heatmap of \textit{``Abalone''} dataset from Stasy-AutoDiff. } 
\end{figure}
\begin{figure}[!b]
  \centering
  \includegraphics[width=0.64\textwidth]{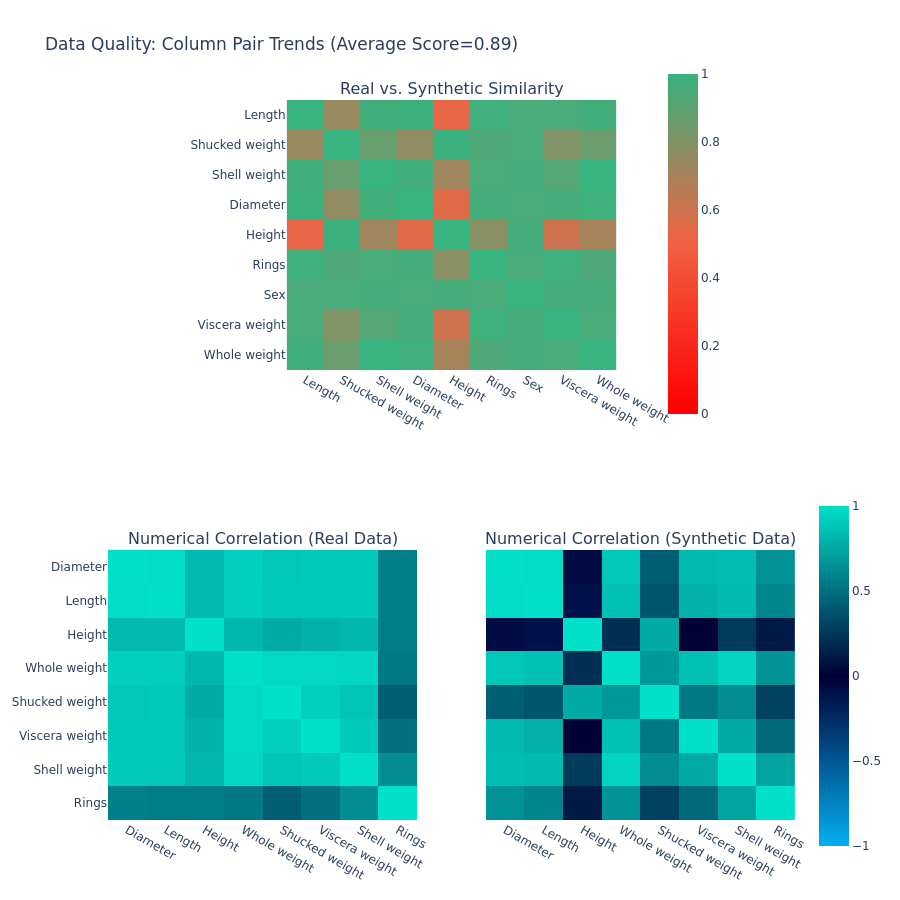}
  \caption{Correaltion Heatmap of \textit{``Abalone''} dataset from TabDDPM. } 
\end{figure}

\newpage
\section{DCR values for each dataset} \label{Appendix4}
We provide the Mean-DCR values averaged over 10 synthetic datasets for each dataset. 
The smallest (resp. second to the smallest) is colored in black (resp. underlined), which implies synthesized datasets are close to the real. 
\begin{table}[!htbp]
\centering
\begin{tabular}{lcccccccc} 
\toprule
 & \textbf{Titanic} & \textbf{adult} & \textbf{HTRU} & \textbf{Magic} & \textbf{Bean} & \textbf{wilt} & \textbf{abalone} \\
\midrule
CTGAN & 5.24 & {109.60} & 10.63 & 29.64 & 9953.02 & 25.62 & 0.19 \\
TVAE & \textbf {2.24} & \textbf{61.10} & 6.93 & \textbf{14.54} & 2364.37 & {12.43} & {\ul 0.08} \\
CTABGAN+ & {4.74} & 247.74 & 16.65 & 24.30 & 3778.49 & 17.35 & 0.16 \\
Stasy & {34.36} & 13150.87 & 94.45 & 82.08 & 55911.20 & 345.85 & 0.56 \\
TabDDPM & 4.63 & 126777.67 & \textbf{3.22} & {\ul 16.94} & \textbf{130.32} & 21.56 & 0.15 \\
AutoGAN & {\ul 2.04} & 720.72 & {5.70} & {24.28} & {645.77} & 12.13 & 0.09 \\
\midrule
Stasy-AutoDiff & {2.89} & 195.92 & {4.53} & 16.95 & {432.73} & \textbf {9.66} & {\ul 0.08} \\
Med-AutoDiff & {2.53} & 282.34 & {3.98} & 18.97 & {634.65} & {\ul 10.14} & {0.12} \\
Tab-AutoDiff & {3.18} & {\ul 107.58} & {\ul 3.77} & 18.37 & {\ul 196.68} & {13.87} & \textbf {0.07} \\
\bottomrule
\end{tabular}
\setlength{\tabcolsep}{3.5pt} 
\begin{adjustwidth}{-3cm}{0cm}
\begin{center}
\begin{longtable}{lccccccccc} 
\toprule
 & \textbf{Patient} & \textbf{nursery} & \textbf{faults} & \textbf{Obesity} & \textbf{News} & \textbf{INSR} & \textbf{CM} & \textbf{SHP} \\
\midrule
CTGAN & 46.25 & {0.28} & 1148213 & 3.79 & {\ul 11587.43} & 35.00 & 693.81 & 30.08 \\
TVAE & \textbf{20.90} & 0.39 & 496487.4 & 3.38 & \textbf {6736.04} & {\ul 29.03} & {\textbf {590.43}} & {\ul 19.38} \\
CTABGAN+ & 106.62 & {\ul 0.23} & 1499634 & 3.10 & 24340.9 & 48.93 & 749.84 & 30.50 \\
Stasy & 81.29 & \textbf{0.22} & 3327054 & 4.86 & 2.71 $\times 10^{8}$ & 1106.14 & 7017.03 & 4824.17 \\
TabDDPM & 256.59 & 0.95 & 4137191 & 98.82 & 705767.7 & 50.29 & 1699.10 & 9316.37 \\
AutoGAN & 46.80 & 1.01 & 300791.7 & 3.47 &  40104.49  & 35.47 & 5851.55 & \textbf{17.10} \\
\midrule
Stasy-AutoDiff & {\ul 33.59} & 0.40 & {\textbf {152907.3}} & \textbf{2.25} & {15147.05} & \textbf {27.32} & {707.86} & 27.83 \\
Med-AutoDiff & {31.11} & 0.56 & {68837.95} & 2.69 & {20369.19} & 30.65 & 896.85 & 28.32 \\
Tab-AutoDiff & {75.77} & \textbf{0.22} & {\ul 278249.9} & {\ul 2.65} & {37245.33} & 38.22 & {\ul 688.58} & 64.98 \\
\bottomrule
\label{tab:mytable}
\end{longtable}
\end{center}
\end{adjustwidth}
\end{table}

\end{document}